\documentclass[runningheads]{llncs}

\usepackage[T1]{fontenc}
\usepackage{graphicx}
\usepackage{booktabs}
\usepackage{multirow}
\usepackage{makecell}
\usepackage{longtable}
\usepackage{placeins}
\usepackage{rotating}
\usepackage{appendix}
\usepackage{graphicx}
\usepackage{subcaption}
\usepackage{comment}
\usepackage{float}
\usepackage{placeins}
\usepackage{dblfloatfix}
\usepackage{array}
\usepackage{url}
\usepackage{lipsum}
\usepackage[a-1b]{pdfx}

\makeatletter
\newcommand{\printfnsymbol}[1]{%
  \textsuperscript{\@fnsymbol{#1}}%
}
\makeatother

\begin{document}

\title{When Adaptation Hurts: Connecting Representational Drift to OOD Failures in MedSAM Fine-Tuning.}
\titlerunning{When Adaptation Hurts}

\author{
Marko Haralović\orcidID{0009-0004-1178-9964}\inst{1,2}
\thanks{Equal contribution.}
\and
Sounic Akkaraju\orcidID{0009-0003-7888-4176}\inst{1}
\thanks{Equal contribution; corresponding author.}
\and
Carlo Baretta\orcidID{0009-0006-6680-1275}\inst{1}
\and
Vasil Zapryanov\orcidID{0009-0002-1441-4039}\inst{1}
\and
Alexia Briassouli\orcidID{0000-0002-0545-3215}\inst{1}
}

\authorrunning{Haralović et al.}

\institute{
University of Zagreb, Faculty of Electrical Engineering and Computing, Zagreb, Croatia\\
\and
University of Twente, Enschede, The Netherlands
\email{
{s.akkaraju,c.d.a.baretta,v.z.zapryanov}@student.utwente.nl;
marko.haralovic@fer.hr;
a.briassouli@utwente.nl
}
}

\maketitle

\begin{abstract}

Foundation models for medical image segmentation, like prompt-based MedSAM, generalize well across domains and modalities, often in zero or few-shot setups. However, their performance depends on the quality of prompts and the adaptation of the models to custom datasets. This work systematically examines how MedSAM generalizes across diverse medical imaging benchmarks, with six adaptation strategies: full-model and encoder-only LoRA, shallow and deep visual prompt tuning (VPT), and decoder-only and full fine-tuning. Models are trained on the International Skin Imaging Collaboration Challenge (ISIC 2018) dataset  and evaluated under clean and increasingly noisy prompts on IN and Out-of-Distribution (OOD) datasets: close-OOD PH2 (dermoscopy), far-OOD BUSI (Breast Ultrasound Images Dataset) and CBIS-DDSM (Curated Breast Imaging Subset of the Digital Database for Screening Mammography). We show that adaptation improves performance on IN and close-OOD data but often reduces performance on far-OOD data. Full fine-tuning provides the best tradeoff, while encoder-only LoRA is the strongest parameter-efficient alternative, outperforming standard LoRA and VPT under far-OOD shifts. Using Centered Kernel Alignment (CKA), we show that far-OOD degradation is strongly associated with drift in decoder representations, whereas encoder similarity alone does not explain robustness. This suggests encoder-only LoRA provides stronger robustness than standard LoRA by adapting the encoder to distribution shift in visual features, while preserving the decoder pathway. We further show that random 0–100 pixel jitter on prompts produces more robust and better performing models. We thus conclude that robust MedSAM adaptation requires the combined consideration of prompt noise exposure, domain shift, and representation preservation. We release our code: \url{https://github.com/ImSounic/medsam-vpt}.


\end{abstract}
\section{Introduction}
\label{sec:introduction}
Semantic segmentation is essential in many clinical applications, but requires costly manual annotation. While deep learning can reduce this burden, task-specific models often fail to generalize beyond their training domains. Segmentation foundation models (FMs) such as SAM~\cite{wang2023sam} offer a prompt based alternative. Out-of-the-box SAM performance in medical imaging faces challenges, such as domain shift, weak boundaries, low contrast, and modality-specific appearance changes~\cite{mazurowski2023segment,wang2024sammed3d}. This has motivated medical segmentation FMs~\cite{cheng2023sam,ma2024medsam,ma2025medsam2} like MedSAM~\cite{ma2024medsam}, which was trained on over one million medical images, across 150 tasks, but assumes region of interest (ROI) prompts at inference time.

The first challenge this approach faces is ROI sensitivity. Promptable models that require ROI prompts~\cite{ma2024medsam,ma2025medsam2} highly depend on the user-provided bounding box, which in clinical settings may be noisy, inconsistent, or unavailable, substantially affecting performance~\cite{cheng2023sam,huang2024roboxsam,yao2024fnpcsam}. Previous work has therefore studied strategies to reduce ROI noise sensitivity, by identifying optimal prompting strategies~\cite{huang2024roboxsam,yao2024fnpcsam} or by introducing perturbed bounding-box prompts~\cite{rahman2024ppsam}. 

The second challenge is generalization under distribution shift, as models lose performance when applied beyond their training distribution~\cite{arega2025posthoc,gutbrod2025openmibood,nguyen2023oodrobustness}. Prior work has studied out-of-distribution (OOD) identification~\cite{arega2025posthoc}, robustness under distribution shift~\cite{nguyen2023oodrobustness,wang2023sam}, and generalization across domains~\cite{cheng2023sam,mazurowski2023segment,nguyen2023oodrobustness}. 

The third challenge is the adaptation cost. Fine-tuning medical FMs can improve performance, but is computationally expensive. Parameter efficient methods such as LoRA~\cite{hu2021lora} and visual prompt tuning (VPT)~\cite{jia2022visual,zhao2025promptseg} reduce this cost, but remain sensitive to dataset and prompt quality~\cite{huang2024roboxsam,mazurowski2023segment,shi2024beyond,yao2024fnpcsam}. As a result, adaptation may improve in-domain performance while degrading robustness to OOD data or noisy ROI prompts. Despite advances, prior work has not systematically connected fine-tuning strategy, prompt robustness, and domain generalization in medical segmentation FMs. Existing studies often focus on target domain adaptation, while separately acknowledging ROI noise and distribution shift. However, it remains unclear which fine-tuning strategies provide the best target-domain performance while preserving generalization, which lose most generalization capabilities, and which are most sensitive to noisy prompts.

\paragraph{\textbf{Contributions}} \label{contributions} This work studies how fine-tuning strategies affect segmentation performance under prompt and domain shift, and relates performance to the internal representations of MedSAM and its fine-tuned variants. In contrast to prior work that focuses mainly on adaptation, prompt refinement, or prompt automation, we jointly evaluate adaptation strategy, prompt robustness, and domain generalization. We compare zero-shot MedSAM with full and decoder-only fine-tuning, full-model and encoder-only LoRA, and shallow and deep VPT under controlled bounding-box perturbations in training and evaluation.

Our contributions are threefold. First, we systematically evaluate bounding-box perturbation, using models trained with clean boxes, fixed-noise boxes, and variable-noise boxes, then evaluated across multiple perturbation levels. Second, we compare full fine-tuning and parameter-efficient fine-tuning in the same perturbation-aware setting to identify which methods remain robust when ROI prompts are imperfect, while test data shifts away from the target domain. Finally, we link layerwise representational similarity, measured with centered kernel alignment (CKA)~\cite{kornblith2019similarity}, to in- and OOD segmentation performance, and test whether representation preservation is associated with OOD generalization.

\section{Related Work}
Recent progress in foundation models has led to segmentation systems that transfer well to medical imaging. MedSAM~\cite{ma2024medsam} is a notable example, adapting SAM~\cite{wang2023sam} to medical images, achieving strong performance across modalities, with extended following versions~\cite{ma2024medsam,ma2025medsam2,wang2024sammed3d,wen2024psam}. Medical datasets are often small and costly to annotate, so foundation models are often evaluated in zero-shot settings. However, out-of-the-box performance is often suboptimal, largely due to distribution shift between pretraining and target clinical data~\cite{cheng2023sam,mazurowski2023segment,wang2023sam}. Fine-tuning on target domain data can improve performance, but it is computationally expensive, requires annotated data, and may reduce generalization ability.

Therefore, lightweight methods have been explored, like prompt tuning~\cite{dutt2024peft}, visual prompt tuning~\cite{CUI2025102608,HE2025107168,jia2022visual,prompttuning_sam_2025}, and low rank adaptation (LoRA) adapters~\cite{dutt2024peft,hu2021lora}. These methods aim to preserve the benefits of large pre-trained models while reducing training cost, but remain sensitive to task design and prompt quality.

\subsection{Efficient fine-tuning methods}

\noindent\textbf{MedSAM architecture} MedSAM~\cite{ma2024medsam} is the backbone to be evaluated under zero-shot and fine-tuned settings. It consists of an image encoder, prompt encoder, and mask decoder. The image encoder is a Vision Transformer (ViT) with 12 transformer blocks that produce dense image embeddings. The prompt encoder embeds user prompts, such as bounding boxes, which the two-layer mask decoder fuses with image embeddings to produce segmentation masks.

\noindent\textbf{Prompt tuning} Prompt tuning freezes most of the network and learns only a small number of task specific parameters, reporting good performance in~\cite{dutt2024peft}.

\noindent\textbf{Visual prompt tuning (VPT)}  VPT extends prompt adaptation to vision models by learning a set of prompts while freezing the backbone \cite{HE2025107168,jia2022visual}, improving performance in a low-data regime~\cite{zhao2025promptseg}, supported by recent SAM variants~\cite{CUI2025102608,prompttuning_sam_2025}.

\noindent\textbf{Low Rank Adaptation (LoRA)} LoRA is a parameter-efficient fine-tuning strategy that updates a model through low-rank trainable weight matrices rather than modifying the full weight tensors, while keeping backbone fixed~\cite{dutt2024peft,hu2021lora}.

\subsection{Robustness and out-of-distribution (OOD) generalization}
Generalization under domain shift remains a challenge in medical segmentation, since models often degrade when transferred across scanners, hospitals, or modalities~\cite{arega2025posthoc,gutbrod2025openmibood}. Related studies have focused on OOD identification~\cite{arega2025posthoc}, robustness under shift~\cite{wang2023sam}, and generalization across domains~\cite{cheng2023sam,mazurowski2023segment}.

\subsection{Prompt sensitivity to noise}

Prior work has shown that promptable segmentation models are sensitive to prompt type and quality~\cite{cheng2023sam,huang2024roboxsam,mazurowski2023segment,wang2024sammed3d,yao2024fnpcsam}, and that perturbing the box can significantly reduce segmentation accuracy, with~\cite{cheng2023sam}, reporting strong variation across datasets and prompt settings~\cite{mazurowski2023segment,wang2024sammed3d}. RobustMedSAM~\cite{li2026robustmedsam} claims image encoder preserves medical priors, while the mask decoder governs corruption robustness. Several works address this limitation by refining or augmenting the input prompt~\cite{huang2024roboxsam,yao2024fnpcsam}, through bounding box correction~\cite{yao2024fnpcsam}, refining noisy prompts~\cite{huang2024roboxsam} or by training on perturbed bounding boxes~\cite{ma2024medsam,rahman2024ppsam}, where imprecise bounding boxes serve as an unsupervised localization task~\cite{shi2024beyond}. 

\subsection{Feature representation comparison}
Comparing representations is a well-studied problem in machine learning~\cite{kornblith2019similarity}. Comparing only final embeddings can miss layer-specific changes, originating from fine tuning strategies that adapt different parts of MedSAM. We therefore compare internal feature representations across layers using Centered Kernel Alignment (CKA)~\cite{kornblith2019similarity}. For each model, we compute CKA between its layer representations and those of zero shot MedSAM, using the same input images.

\section{Methodology}
\label{sec:methodology}

\subsection{Perturbed bounding box prompts}
\label{subsec:perturbed_bbox_prompts}

All images are resized to $1024 \times 1024$. The bounding box prompts undergo evaluation perturbations of $p \in \{0,20,50,100,200\}$, corresponding to mean side length increases of $0\%$, $1.95\%$, $4.88\%$, $9.77\%$, and $19.53\%$ of image size, and mean box area increases of $0\%$, $17.61\%$, $48.85\%$, $112.67\%$, and $280.14\%$, reaching as high as $1327.9\%$ for small lesions in CBIS-DDSM.

\textit{During training:}
\label{par:bbox_perturbation_training} 
We train models under three prompt settings: clean bounding boxes, bounding boxes perturbed by a fixed 20 pixels, and bounding boxes randomly perturbed between 0 and 100 pixels. This allows us to test whether exposure to imperfect prompts improves localization robustness, and whether training with a fixed perturbation causes overfitting to a specific noise level.

\textit{During evaluation}
\label{par:bbox_perturbation_evaluation}
We evaluate each trained model across five bounding box perturbation levels: 0, 20, 50, 100, and 200 pixels. This tests robustness under increasing prompt noise and evaluates whether models trained with perturbations up to 100 pixels generalize to more severe 200 pixel perturbations in inference.

\begin{figure}[!t]
    \centering
    \includegraphics[width=\linewidth,keepaspectratio]{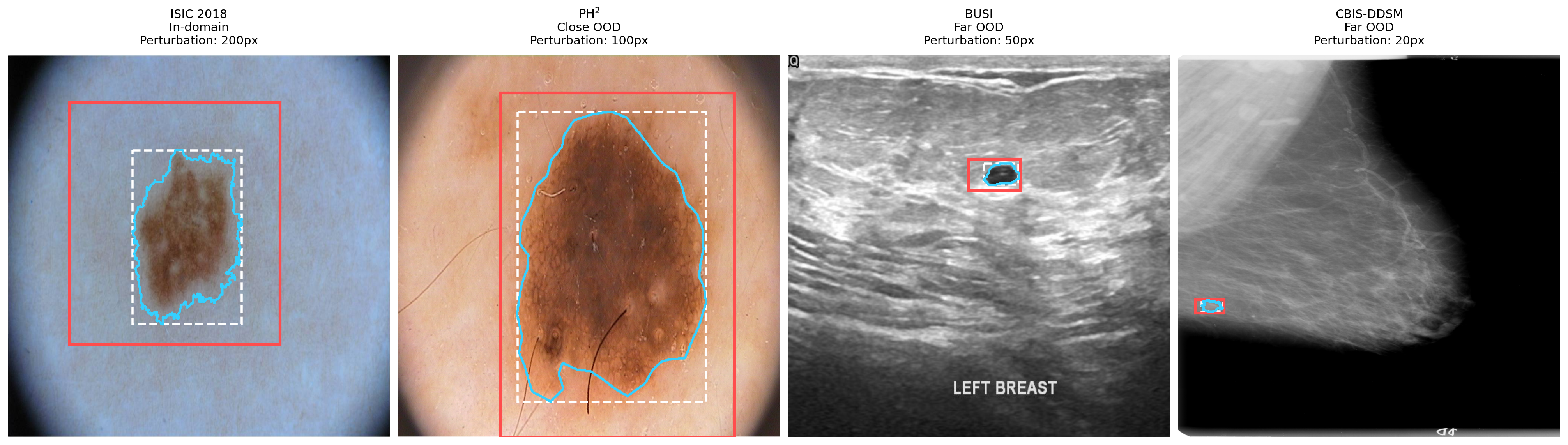}
    \caption{Prompt perturbation examples.}
    \label{fig:prompt_perturbation_examples}
\end{figure}

\subsection{Datasets and in- vs out-of-domain separation}
\label{sec:datasets_domain_split}

We use ISIC 2018~\cite{codella2019isic2018} dermoscopic skin lesion images [CC BY-NC] for training and in-domain evaluation. PH$^2$~\cite{mendonca2013ph2} [public research use], another dermoscopic skin lesion dataset, is used as a close out-of-domain dataset because it shares the same imaging modality and target structure but differs in acquisition source and dataset distribution. To evaluate stronger domain shift, we use BUSI breast ultrasound images~\cite{al2020busi} [CC BY 4.0] and CBIS-DDSM mammography images~\cite{lee2017cbisddsm} [TCIA data usage policy] as far out-of-domain datasets. This separation allows testing whether adaptation improves performance only for target domain or preserves generalization across different clinical imaging domains.

\subsection{Representation similarity measurement}
\label{subsubsec:representation_similarity}

We measure representation similarity using linear Centered Kernel Alignment (CKA). For each adapted checkpoint, we compare layerwise representations against zero shot MedSAM using the same images and prompts. CKA is computed across the image encoder, prompt encoder, and mask decoder, with higher values indicating stronger preservation of the original MedSAM representation. 
\section{Experiments and results}
\label{sec:experiments}

\subsection{Implementation details}

\textbf{Training paradigms implementation}
All methods use the MedSAM backbone with a frozen prompt encoder, so bounding-box prompts are encoded identically across experiments. Full fine-tuning updates the image encoder and mask decoder, while decoder-only fine-tuning updates only the mask decoder. Full LoRA adds trainable adapters to both the image encoder and decoder, whereas encoder-only LoRA adds adapters only to the image encoder. VPT freezes the image encoder and adds learnable visual prompt tokens either once at the encoder input for shallow VPT or at every encoder block for deep VPT; in both VPT settings, the mask decoder is trained.

\textbf{Training details} All models were trained for 5 epochs on ISIC 2018 at \(1024 \times 1024\) resolution using AdamW, mixed precision, and an equally weighted Dice + cross-entropy loss. The 2,594 ISIC images were split 80/10/10 into train, validation, and test sets. Experiments used 8 GPUs with 16--22 GB memory, while training lasted 8 hours. Full fine-tuning used batch size 1 and learning rate \(10^{-5}\), while parameter-efficient methods used batch size 4 and learning rate \(10^{-4}\). Hyperparameters were fixed across datasets, Dice was reported for readability, and each configuration was repeated across three random seeds. For LoRA, we used rank (r=8), scaling factor ($\alpha=16$), and dropout (0.0); for VPT, we used 10 prompt tokens per insertion point. The number of trainable parameters was 93.7M for full fine-tuning, 4.4M for full LoRA, 4.1M for encoder-only LoRA, 4.1M for decoder-only fine-tuning, and 4.1M/4.2M for shallow/deep VPT. CKA and OOD evaluation were performed on 200 PH2, 647 BUSI, and 362 CBIS-DDSM images. For ISIC, CKA was computed on all 2,594 images.

\subsection{Per-method performance across variable evaluation jitter}
\label{sec:per_method_performance}

\begin{figure}[!t]
    \centering
    \includegraphics[width=0.8\linewidth]{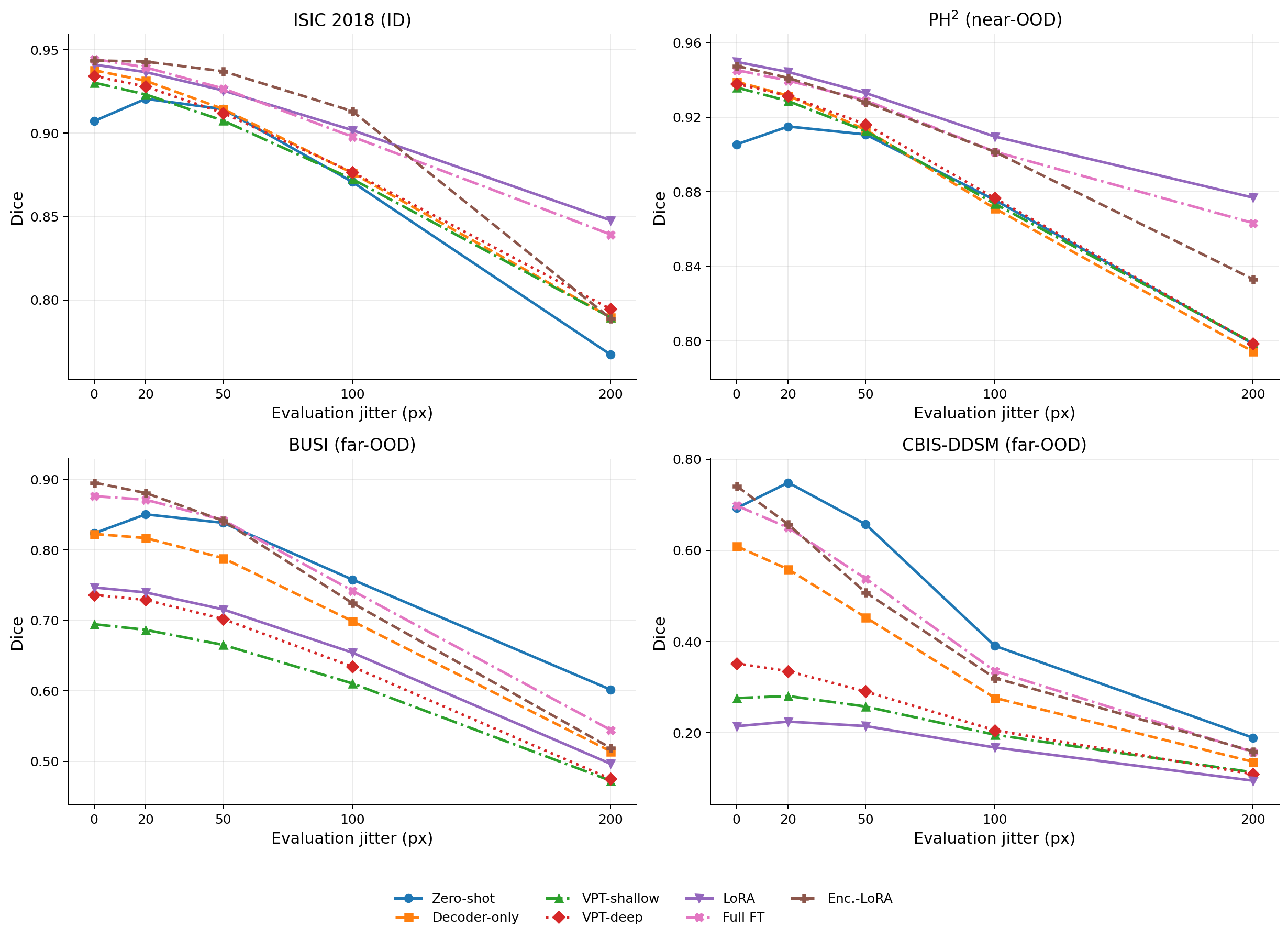}
    \caption{
    Robustness to bounding-box perturbation under variable training jitter.
    Dice is reported across evaluation jitter levels for all adaptation strategies on ID, near-OOD, and far-OOD datasets.
    }
    \label{fig:robustness_curves}
\end{figure}

Figure~\ref{fig:robustness_curves} summarizes the robustness of each method across jitter levels. For each method, we average performance across the three training prompt perturbations: 0, fixed 20, and random 0--100 pixels. Overall, performance degrades substantially as evaluation jitter increases. The results also show variability across adaptation methods. Full fine-tuning and encoder-only LoRA are the strongest and most robust methods overall. Standard LoRA, drops substantially on OOD datasets, despite strong in-domain performance, as reported in the literature~\cite{arega2025posthoc}. At higher jitter levels, zero-shot outperforms all adapted models on BUSI and CBIS-DDSM. Only full fine-tuning and encoder-LoRA maintain far OOD performance close to zero-shot while still improving in-domain performance.

\subsection{Performance regarding method and training jitter}
\label{sec:cka_corr_w_ood}

Table~\ref{tab:regime_gain_by_training_compact} summarizes Dice gain relative to zero-shot MedSAM across adaptation methods and training jitter. Overall, zero-shot performance is difficult to improve upon, especially on far-OOD datasets, while in-domain and close-OOD gains often come with reduced generalization.

Standard LoRA achieves the largest ID and close-OOD gains, but drops sharply under far-OOD shift, with a mean gain of -0.176. In contrast, encoder-only LoRA is more robust, with a far-OOD mean of -0.043 and the second-best overall mean of -0.016. Better generalization by restricting LoRA to the encoder is consistent with the CKA analysis (Section~\ref{sec:cka_corr_w_ood}) that shows preserving decoder representations is associated with far-OOD robustness. Visual prompt tuning does not provide a favorable tradeoff: shallow and deep VPT show negligible ID and close-OOD gains, with large far-OOD drops. Training with random 0--100-pixel jitter generally produces the best adapted models overall, especially for full fine-tuning and encoder-only LoRA.

\begin{table}[!t]
\centering
\caption{Mean Dice gain over zero-shot by regime, method, and training jitter. Values are averaged over three runs; the Mean column reports mean $\pm$ standard deviation. \textbf{Bold} and \underline{underline} indicate the best and second-best methods.}
\label{tab:regime_gain_by_training_compact}
\setlength{\tabcolsep}{3pt}
\renewcommand{\arraystretch}{0.7}
\scriptsize
\begin{tabular}{@{}llrrrrcc@{}}
\toprule
Regime & Method & 0 & 20 & r100 & Mean & Best train & Best method \\
\midrule
\multirow{6}{*}{ID}
& Dec. only & -0.016 & -0.007 & \phantom{-}0.014 & \(-0.003 \pm 0.002\) & r100 &  \\
& VPT-S     & -0.008 & -0.001 & \phantom{-}0.009 & \(-0.000 \pm 0.003\) & r100 &  \\
& VPT-D     & -0.010 & -0.005 & \phantom{-}0.013 & \(-0.001 \pm 0.002\) & r100 &  \\
& LoRA      & \phantom{-}0.012 & \phantom{-}0.016 & \phantom{-}0.035 & \(\mathbf{0.021 \pm 0.001}\) & r100 & \textbf{LoRA} \\
& Enc-LoRA  & -0.027 & -0.028 & \phantom{-}0.053 & \(-0.001 \pm 0.002\) & r100 &  \\
& Full FT   & -0.006 & -0.005 & \phantom{-}0.034 & \(\underline{0.007 \pm 0.002}\) & r100 &  \\
\midrule
\multirow{6}{*}{Close OOD}
& Dec. only & -0.016 & -0.007 & \phantom{-}0.009 & \(-0.005 \pm 0.004\) & r100 &  \\
& VPT-S     & -0.009 & -0.004 & \phantom{-}0.009 & \(-0.001 \pm 0.003\) & r100 &  \\
& VPT-D     & -0.011 & -0.007 & \phantom{-}0.011 & \(-0.002 \pm 0.001\) & r100 &  \\
& LoRA      & \phantom{-}0.015 & \phantom{-}0.020 & \phantom{-}0.042 & \(\mathbf{0.026 \pm 0.001}\) & r100 & \textbf{LoRA} \\
& Enc-LoRA  & -0.017 & -0.022 & \phantom{-}0.023 & \(-0.005 \pm 0.002\) & r100 &  \\
& Full FT   & -0.007 & -0.003 & \phantom{-}0.035 & \(\underline{0.008 \pm 0.002}\) & r100 &  \\
\midrule
\multirow{6}{*}{Far OOD}
& Dec. only & -0.036 & -0.032 & -0.088 & \(-0.052 \pm 0.010\) & 20 &  \\
& VPT-S     & -0.127 & -0.161 & -0.230 & \(-0.172 \pm 0.002\) & 0 &  \\
& VPT-D     & -0.126 & -0.139 & -0.198 & \(-0.154 \pm 0.015\) & 0 &  \\
& LoRA      & -0.142 & -0.157 & -0.228 & \(-0.176 \pm 0.017\) & 0 &  \\
& Enc-LoRA  & -0.024 & -0.052 & -0.054 & \(\underline{-0.043 \pm 0.010}\) & 0 &  \\
& Full FT   & -0.014 & -0.011 & -0.029 & \(\mathbf{-0.018 \pm 0.005}\) & 20 & \textbf{Full FT} \\
\midrule
\multirow{6}{*}{Overall}
& Dec. only & -0.023 & -0.015 & -0.022 & \(-0.020 \pm 0.005\) & 20 &  \\
& VPT-S     & -0.048 & -0.055 & -0.071 & \(-0.058 \pm 0.003\) & 0 &  \\
& VPT-D     & -0.049 & -0.050 & -0.058 & \(-0.052 \pm 0.004\) & 0 &  \\
& LoRA      & -0.038 & -0.040 & -0.051 & \(-0.043 \pm 0.006\) & 0 &  \\
& Enc-LoRA  & -0.023 & -0.034 & \phantom{-}0.007 & \(\underline{-0.016 \pm 0.004}\) & r100 &  \\
& Full FT   & -0.009 & -0.006 & \phantom{-}0.013 & \(\mathbf{-0.001 \pm 0.003}\) & r100 & \textbf{Full FT} \\
\bottomrule
\end{tabular}
\end{table}

Across all regimes, full fine-tuning remains the strongest method, achieving the best far-OOD mean gain (-0.018) and best overall mean gain (-0.001). Encoder-only LoRA is the strongest parameter-efficient alternative, clearly outperforming standard LoRA and VPT under OOD shift.

\subsection{CKA correlation with OOD performance}
\label{sec:cka_corr_w_ood}

The following CKA analysis links performance to decoder representational drift, as each training paradigm inevitably shifts internal model representations across layers. Here, we connect which MedSAM internal representations correlate with performance gains. Table~\ref{tab:cka_regime_corr} shows that CKA is weakly correlated with performance gains in ID and close-OOD regimes, but strongly correlated with far-OOD gains for decoder-side representations. The IoU token output, output layer, decoder layers, and upscaled embedding show high positive far-OOD correlations. In contrast, encoder-layer CKA is weakly negative across regimes, indicating that encoder similarity to zero-shot MedSAM alone does not explain robustness.

Overall, far-OOD degradation is tied to drift in the mask decoder and output representations, suggesting that generalization is harmed when output-side representations are disrupted. Therefore we include encoder-only LoRA fine-tuning, which performs better both overall and on far-OOD data than standard LoRA, while remaining competitive with full FT. This analysis, often ignored by works focusing mainly on Dice optimization, suggests encoder adaptation of MedSAM can result in a much more robust model overall.

Because the pooled CKA observations share images and backbones across methods, seeds, and perturbation levels, they are not independent and the effective sample size is smaller than the nominal $n$. The reported $p$-values should therefore be interpreted as descriptive rather than strict hypothesis tests. This does not change the qualitative conclusion that decoder-side similarity tracks far-OOD performance more closely than encoder similarity.

\begin{table}[!t]
\centering
\caption{Spearman/Pearson correlations between CKA and Dice gain over zero-shot. Stars denote false discovery rate corrected significance across 30 tests: $^*$ $q<0.05$, $^{**}$ $q<0.01$, $^{***}$ $q<0.001$. Total sample size is $n=3000$.}
\label{tab:cka_regime_corr}
\setlength{\tabcolsep}{4pt}
\footnotesize
\begin{tabular}{lccc}
\toprule
CKA feature & ID (ISIC 2018) & Close-OOD (PH$^2$) & Far-OOD (BUSI/CBIS) \\
\midrule
IoU token layer   &  0.097 /  0.106 & -0.018 / -0.021 & 0.760$^{***}$ / 0.692$^{***}$ \\
Output layer       &  0.060 /  0.026 &  0.004 / -0.017 & 0.755$^{***}$ / 0.718$^{***}$ \\
Decoder layers     & -0.075 / -0.107 & -0.022 / -0.079 & 0.726$^{***}$ / 0.687$^{***}$ \\
Upscaled emb.      &  0.195 /  0.050 &  0.294$^{*}$ /  0.213 & 0.713$^{***}$ / 0.791$^{***}$ \\
Encoder layers     & -0.282$^{*}$ / -0.299$^{*}$ & -0.281$^{*}$ / -0.352$^{**}$ & 0.091 / 0.059 \\
\bottomrule
\end{tabular}
\end{table}
\section{Conclusion}
We systematically evaluated MedSAM adaptation under joint domain shift and bounding-box prompt perturbation, both during training and evaluation. By training with clean boxes, fixed 20-pixel, and random 0--100 pixel perturbations, and evaluating across increasing prompt noise, we show that prompt robustness depends strongly on both the adaptation strategy and the training jitter regime. Zero-shot MedSAM remains difficult to improve upon when far-OOD robustness is considered, especially under severe prompt perturbation.

Across adaptation methods, full fine-tuning provides the strongest overall tradeoff, while encoder-only LoRA is the best parameter-efficient alternative. Compared with standard LoRA, decoder-only fine-tuning, and shallow or deep visual prompt tuning, encoder-only LoRA better preserves far-OOD performance while maintaining strong in-domain and close-OOD performance.

Our CKA analysis links performance differences to decoder representational drift. Far-OOD generalization is strongly associated with preserving decoder and output representations, whereas encoder similarity does not explain robustness. This supports encoder-only LoRA as a practical CKA-informed adaptation strategy: adapting the image encoder while avoiding decoder/output drift leads to a stronger robustness tradeoff than standard LoRA. Overall, variable 0--100 pixel jitter training produces the most reliable adapted models.

\noindent\textbf{Disclosure of Interests.} The authors have no competing interests in the paper.

\bibliographystyle{splncs04}
\bibliography{references}

\newcommand{\maketitlesupplementary}{
\begin{center}
{\Large \bfseries When Adaptation Hurts: Connecting Representational Drift to OOD Failures in MedSAM Fine-Tuning.\par}
\vspace{1.0em}
{\Large \bfseries Supplementary Material\par}
\end{center}
\vspace{1.5em}
}

\clearpage
\appendix
\maketitlesupplementary
\FloatBarrier

\section{How hard of a task does each evaluation jitter represent?}

In Table~\ref{tab:bbox_jitter_relative_area} we report that the same pixel-level jitter corresponds to very different effective prompt difficulty across datasets. For ISIC 2018 and PH$^2$, the increase in bounding-box area remains relatively moderate even at larger perturbations, whereas BUSI and especially CBIS-DDSM experience much larger relative ROI expansions, reflecting the smaller lesion sizes and more severe prompt degradation in those domains. This means that a fixed jitter value does not represent the same level of challenge across all datasets, and that far-OOD evaluation is substantially more demanding than in-domain evaluation. Taken together, these results support our use of jitter-aware evaluation as a necessary component for assessing robustness under noisy ROI prompts.

\begin{table}[H]
\centering
\small
\setlength{\tabcolsep}{4pt}
\caption{
Relative increase in tight bounding-box area under expansion due to evaluation jitter. Values report the mean area increase with respect to the original tight bounding box for each dataset and perturbation level.
}
\label{tab:bbox_jitter_relative_area}
\begin{tabular}{lcccc}
\toprule
\textbf{Dataset} & \textbf{pm=20} & \textbf{pm=50} & \textbf{pm=100} & \textbf{pm=200} \\
\midrule
ISIC2018-test & $11.70\%$ & $30.51\%$ & $66.40\%$ & $152.26\%$ \\
PH2           & $6.57\%$  & $16.84\%$ & $35.11\%$ & $72.06\%$ \\
BUSI          & $18.71\%$ & $51.36\%$ & $113.75\%$ & $270.96\%$ \\
CBIS-DDSM     & $64.06\%$ & $193.49\%$ & $485.19\%$ & $1327.90\%$ \\
\midrule
\textbf{Overall} & $\mathbf{17.61\%}$ & $\mathbf{48.85\%}$ & $\mathbf{112.67\%}$ & $\mathbf{280.14\%}$ \\
\bottomrule
\end{tabular}
\end{table}
\section{Boundary robustness under evaluation jitter}
\label{app:boundary_robustness}

Table~\ref{tab:boundary_hd95} reports boundary accuracy for clean bounding-box prompts. The adapted models obtain low median HD95 values on ISIC 2018 and PH$^2$. On the far-OOD datasets, the differences between methods are larger. Encoder-only LoRA, decoder-only fine-tuning, and full fine-tuning score relatively low boundary errors, while standard LoRA and the VPT variants have much wider
interquartile ranges, particularly on BUSI and CBIS-DDSM.

\begin{figure}[!t]
    \centering
    \includegraphics[width=0.8\linewidth]{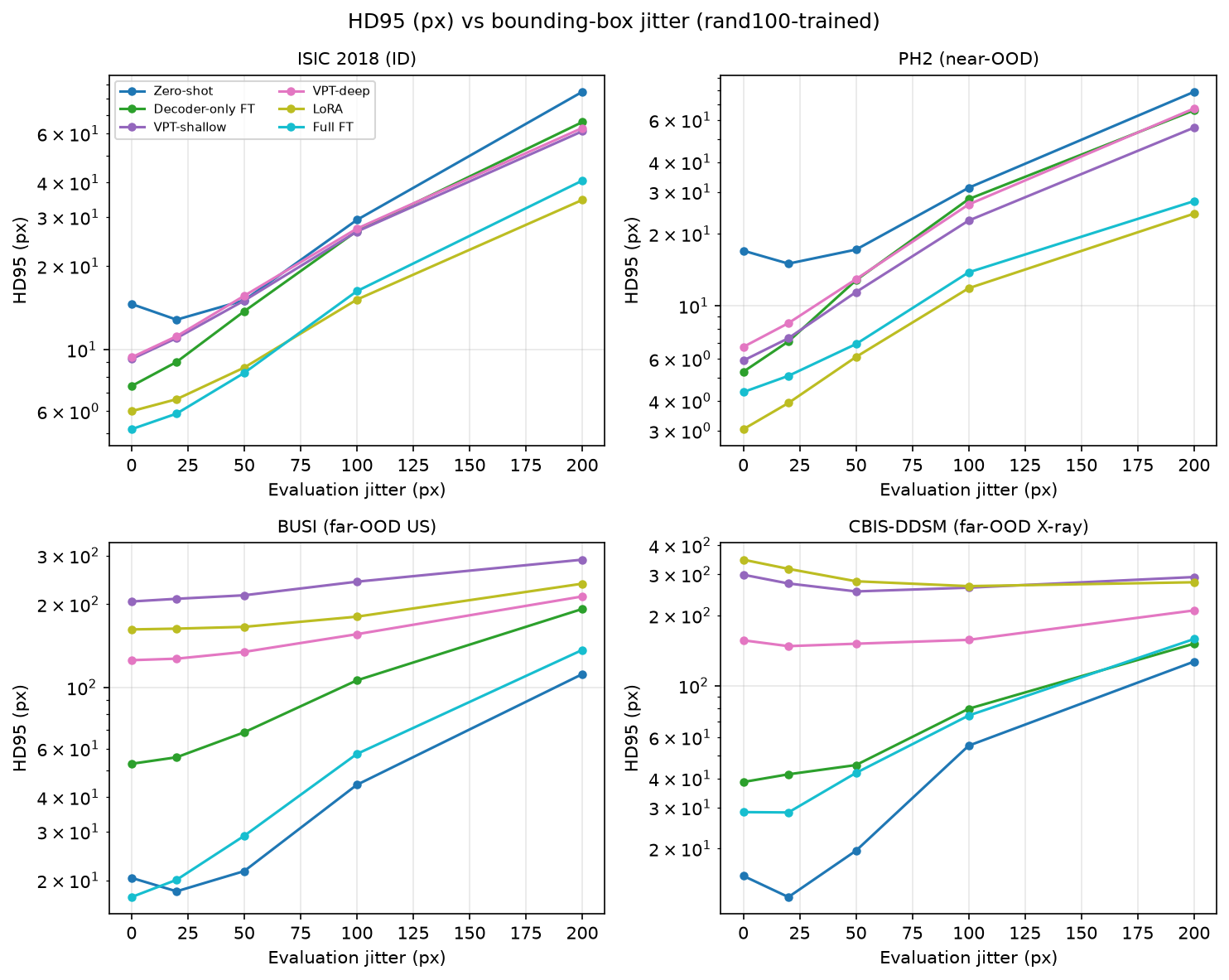}
    \caption{HD95 across evaluation jitter levels for each adaptation method and dataset. Values are shown on a logarithmic scale in pixels at
    $1024\times1024$; lower is better.}
    \label{fig:hd95_robustness}
\end{figure}

Figure~\ref{fig:hd95_robustness} extends this comparison across evaluation jitter levels. Boundary error generally increases as the bounding box is expanded, as the quality of the prompts degrades, but the rate of degradation differs across adaptation methods and datasets. The largest errors occur for standard LoRA and VPT on the far-OOD datasets.

\begin{table}[!t]
\centering
\caption{HD95 at zero evaluation jitter. Values are median [interquartile range] in pixels at $1024\times1024$; lower is better.}
\label{tab:boundary_hd95}
\setlength{\tabcolsep}{3pt}
\renewcommand{\arraystretch}{1.1}
\scriptsize
\begin{tabular}{@{}lcccc@{}}
\toprule
Method &
\shortstack{ISIC 2018\\(ID)} &
\shortstack{PH$^2$\\(near-OOD)} &
\shortstack{BUSI\\(far-OOD)} &
\shortstack{CBIS-DDSM\\(far-OOD)} \\
\midrule
Zero-shot & 6 [0, 15] & 7 [1, 16] & 14 [9, 25]  & 12 [7, 17]   \\
Dec. only & 0 [0, 3]  & 0 [0, 3]  & 4 [2, 17]   & 4 [3, 6]     \\
VPT-S     & 0 [0, 3]  & 0 [0, 4]  & 21 [5, 387] & 14 [8, 37]   \\
VPT-D     & 0 [0, 3]  & 0 [0, 4]  & 25 [6, 394] & 18 [10, 359] \\
LoRA      & 0 [0, 1]  & 0 [0, 1]  & 26 [6, 402] & 47 [13, 438] \\
Enc-LoRA  & 0 [0, 0]  & 0 [0, 0]  & 3 [1, 11]   & 5 [3, 7]     \\
Full FT   & 0 [0, 0]  & 0 [0, 0]  & 4 [1, 13]   & 4 [3, 6]     \\
\bottomrule
\end{tabular}
\end{table}

\section{How does CKA similarity correspond to segmentation performance}

\begin{figure}[H]
    \centering
    \includegraphics[width=\columnwidth]{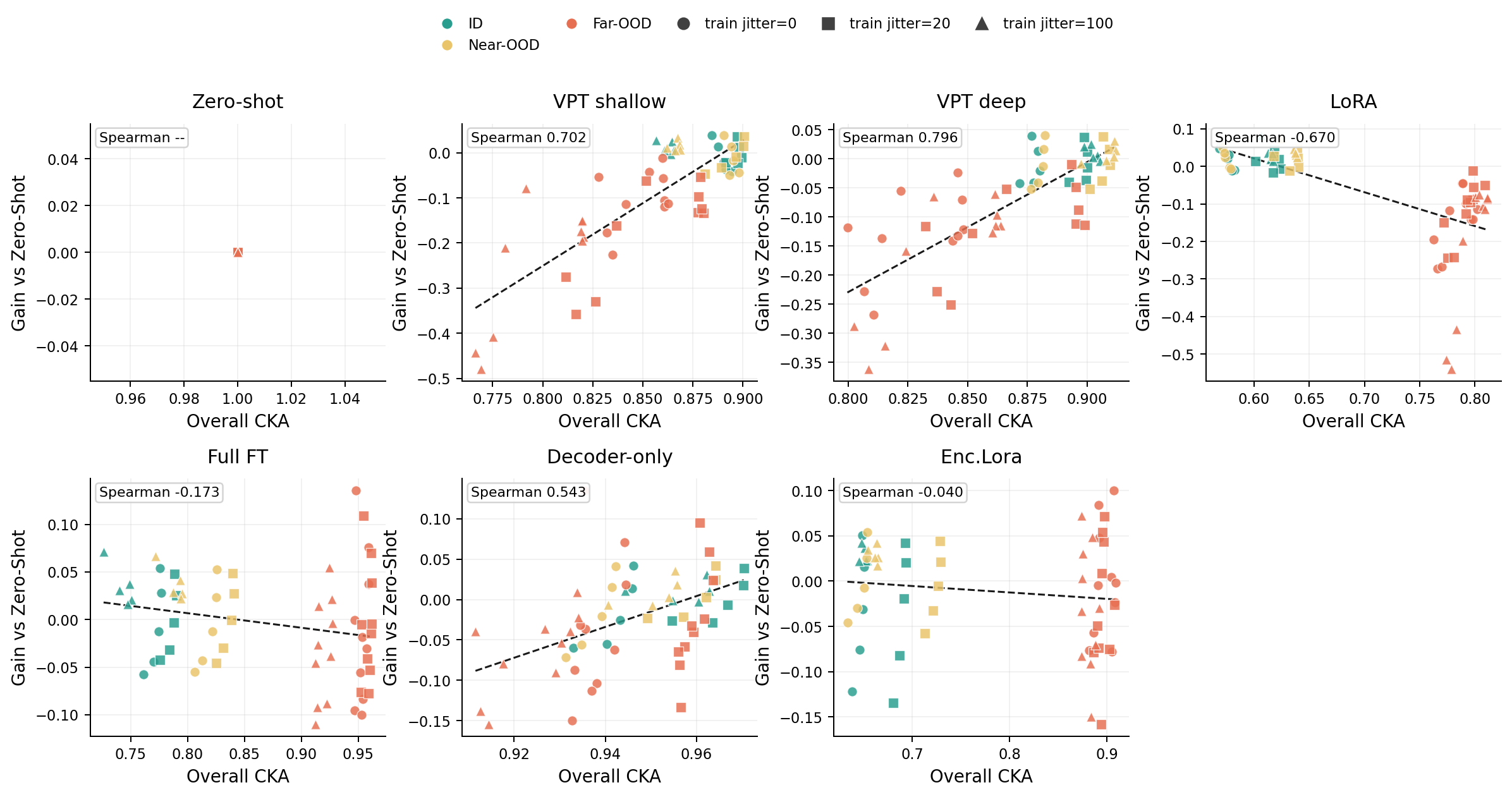}
    \caption{CKA-performance relationship by method.}
    \label{fig:cka_perf_by_method}
\end{figure}

\begin{figure}[H]
    \centering
    \includegraphics[width=\columnwidth]{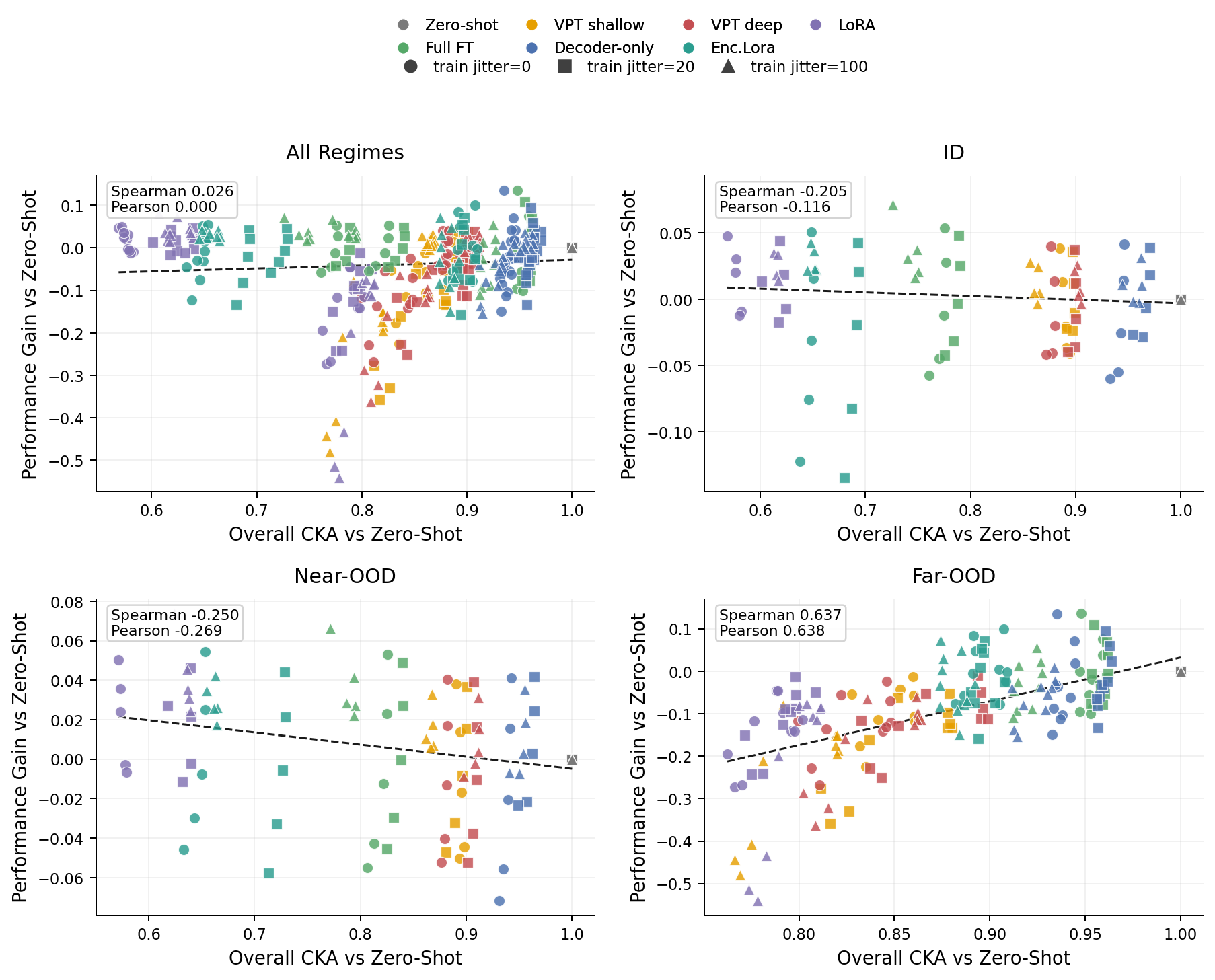}
    \caption{Overview of the relationship between CKA similarity and segmentation performance.}
    \label{fig:cka_perf_overview_scatter}
\end{figure}

Figures~\ref{fig:cka_perf_by_method} and~\ref{fig:cka_perf_overview_scatter} show that CKA similarity is related to segmentation performance, but the relationship depends strongly on which part of the network is compared. The scatter plots indicate that higher similarity in decoder-side and output-side representations is associated with better far-OOD performance, while encoder similarity alone does not track robustness as reliably. The heatmap makes the same pattern easier to see across datasets and jitter levels, where the strongest correlations appear in the far-OOD regime. Overall, these figures support the claim that preserving decoder and output representations is more important for robustness than preserving encoder features alone.

Figure~\ref{fig:cka_perf_by_method} plots the performance gain over zero-shot against overall CKA similarity for all settings. Each method has it's own plot and each point in the scatter plot corresponds to one configuration of the domains (ID, near-OOD, and far-OOD) and training jitter for a given adaptation method. The figure displays how the CKA–performance relationship differs across adaptation strategies while still aggregating over all domain and jitter settings.

\newpage

Figure~\ref{fig:cka_perf_overview_scatter} shows a plot per regime, where each plot shows the performance gain over zero-shot against overall CKA similarity for all settings. Each point in the scatter corresponds to one configuration of the adaptation method, training jitter, and evaluation jitter. Per method, the plots cover four datasets, three training jitters, and five evaluation jitters per method. The results shown in the four plots show how the relationship changes when moving from in-domain to far-OOD conditions.

\begin{figure}[H]
    \centering
    \includegraphics[width=\columnwidth]{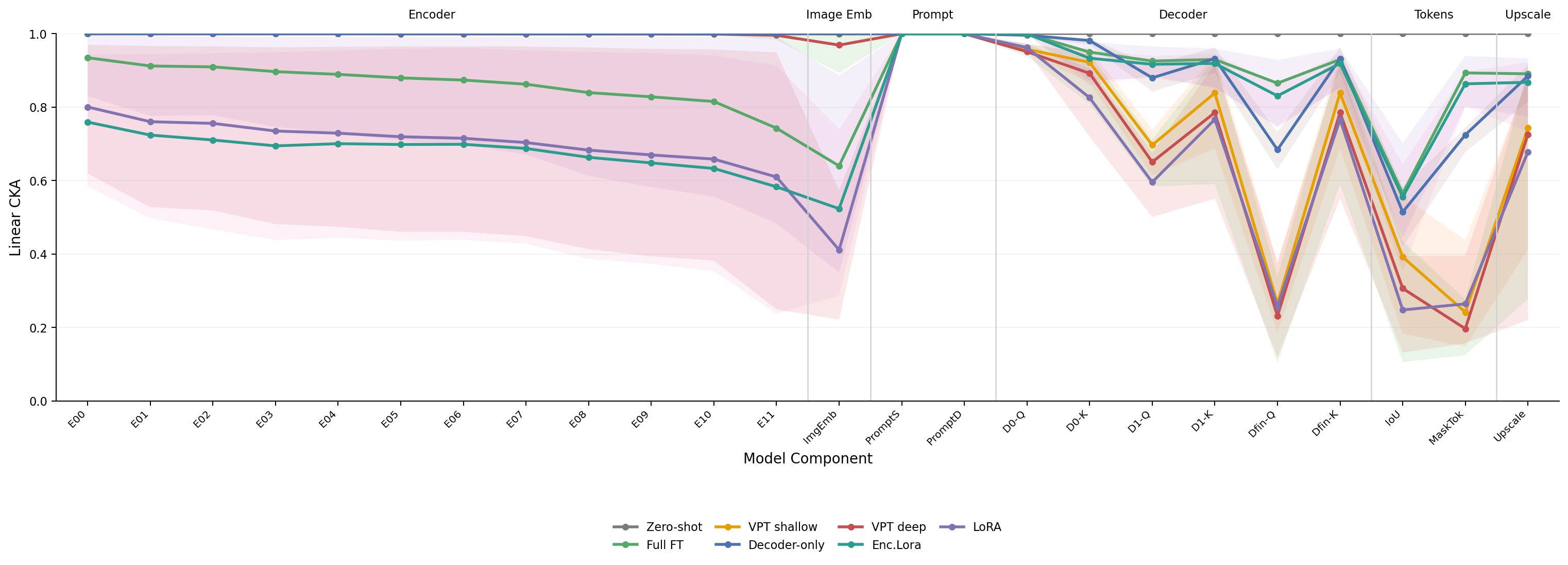}
    \caption{CKA similarity across datasets and jitters using perfect bounding boxes.}
    \label{fig:cka-perfect-pm000-global}
\end{figure}

Figure~\ref{fig:cka-perfect-pm000-global} shows a line plot split into 6 parts. The plot is split into encoder, image emb, prompt, decoder, tokens, and upscale sections. These are the different modules of the model, and the x-axis shows the different components within those modules that we fine-tune. The y-axis shows the linear CKA similarity values. Each line section represents a training method and shows how these different methods alter the internal state of a model compared to its original zero-shot state. The shaded regions around the curves denote variability across datasets and jitters. By showing how fine-tuning affects the similarity scores per method, per component of the model, this plot shows which components are stable and which are sensitive to the fine-tuning task. \\
In the encoder (E00–E11), most adaptation strategies largely preserve the representation of the reference MedSAM model. The zero-shot model, by definition, remains at a similarity of 1.0, while adapted variants show modest but method-dependent drift, with approaches such as LoRA and encoder-only LoRA showing the strongest deviations. \\
In contrast, similarity drops much more sharply in the decoder and token components, particularly for VPT-shallow, VPT deep, and decoder-only tuning, indicating that these parts of the network undergo the largest representational changes during adaptation. The token outputs are substantially reshaped across methods, whereas LoRA variants and full fine-tuning tend to better preserve decoder-side structure, which is consistent with their stronger far-OOD robustness in the main experiments.

\section{Degradation of performance depending on bounding box jitter}
\label{sec:degration_w_jitter}

\begin{figure}[H]
    \centering
    \includegraphics[width=\columnwidth]{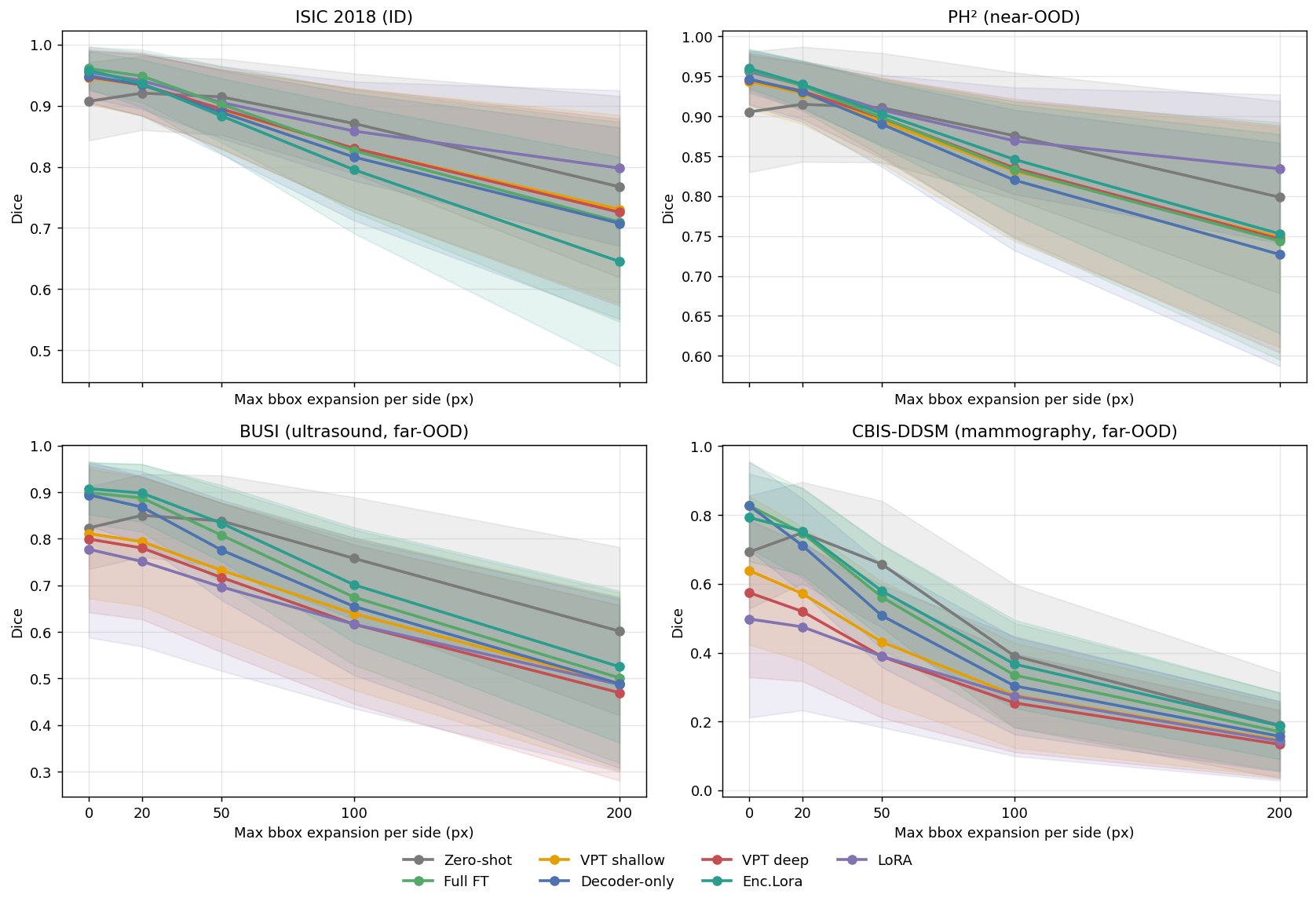}
    \caption{Dice degradation with increasing bounding-box jitter by dataset and method.}
    \label{fig:degradation_curves}
\end{figure}

\begin{figure}[H]
    \centering
    \includegraphics[width=\columnwidth]{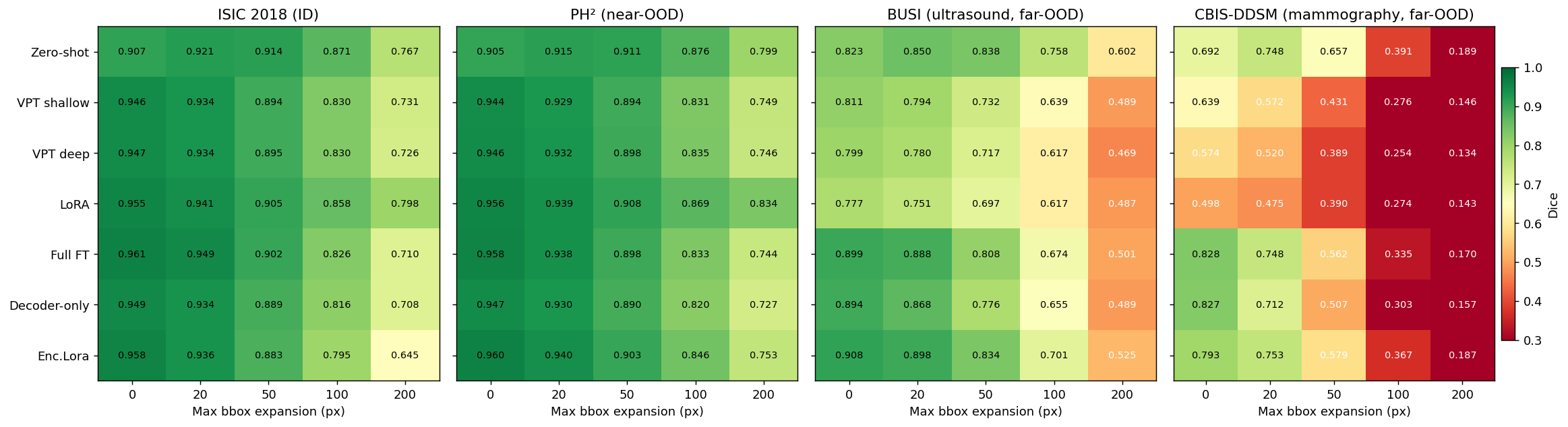}
    \caption{Heatmap of the degradation in Dice with increasing bounding-box jitter, by datasets across models}
    \label{fig:degradation_heatmap}
\end{figure}

Figures~\ref{fig:degradation_curves} and~\ref{fig:degradation_heatmap} show that all methods degrade as bounding-box jitter increases, but they do not degrade at the same rate. Full fine-tuning and encoder-only LoRA have the shallowest degradation curves, especially when they are trained with variable jitter, while standard LoRA and VPT drop more quickly under stronger perturbation. The heatmap makes the same trend visible across datasets, showing that far-OOD performance is much more sensitive to prompt noise than in-domain performance. Specifically for CBIS-DDSM the Dice scores collapse to values below 0.2 for all methods when heavy jitter is introduced to the bounding box dimensions. These results support our claim that exposure to variable prompt noise during training improves robustness under realistic prompt perturbations.

\section{Segmentation performance breakdown across jitter levels}

{
\small
\begin{longtable}{lllcccccc}
\caption{Performance and CKA under bounding-box jitter. Dice performance and CKA are reported at zero jitter, averaged across evaluation jitters, and at 200 px jitter. For performance columns, underline marks the best method within each dataset and training-jitter block, while bold marks the best value for each dataset and evaluation column across all training-jitter settings. For CKA columns, bold and underline mark the best and second-best adapted methods, respectively, within each dataset and training-jitter block}
\label{tab:perf_cka_jitter_ranked}\\
\toprule
Dataset & Training & Method & Dice@0 & mDice & Dice@200 & CKA@0 & mCKA & CKA@200 \\
\midrule
\multirow{22}{*}{\makecell{ISIC 2018 \\ \textit{ID}}} & 0 & Zero-shot & 0.9072 & \underline{0.9019} & 0.7673 & 1.0000 & 1.0000 & 1.0000 \\
 &  & Decoder-only & 0.9487 & 0.8798 & 0.7075 & \textbf{0.9462} & \textbf{0.9432} & \textbf{0.9330} \\
 &  & VPT-shallow & 0.9457 & 0.8858 & 0.7306 & \underline{0.8845} & \underline{0.8908} & \underline{0.8913} \\
 &  & VPT-deep & 0.9472 & 0.8863 & 0.7256 & 0.8767 & 0.8791 & 0.8718 \\
 &  & LoRA & 0.9545 & 0.9015 & \underline{0.7978} & 0.5685 & 0.5794 & 0.5771 \\
 &  & LoRA Encoder-only & 0.9580 & 0.8435 & 0.6450 & 0.6484 & 0.6462 & 0.6379 \\
 &  & Full FT & 0.9609 & 0.8924 & 0.7098 & 0.7755 & 0.7738 & 0.7609 \\
\cmidrule(lr){2-9}
 & 20 & Zero-shot & 0.9072 & \underline{0.9019} & 0.7673 & 1.0000 & 1.0000 & 1.0000 \\
 &  & Decoder-only & 0.9462 & 0.8961 & 0.7408 & \textbf{0.9703} & \textbf{0.9651} & \textbf{0.9546} \\
 &  & VPT-shallow & 0.9429 & 0.8948 & 0.7455 & 0.8969 & 0.8965 & 0.8914 \\
 &  & VPT-deep & 0.9444 & 0.8890 & 0.7277 & \underline{0.8987} & \underline{0.8981} & \underline{0.8924} \\
 &  & LoRA & 0.9510 & 0.8998 & \underline{0.7808} & 0.6184 & 0.6168 & 0.6012 \\
 &  & LoRA Encoder-only & 0.9496 & 0.8414 & 0.6330 & 0.6925 & 0.6888 & 0.6802 \\
 &  & Full FT & \underline{0.9553} & 0.8987 & 0.7252 & 0.7886 & 0.7851 & 0.7754 \\
\cmidrule(lr){2-9}
 & rand100 & Zero-shot & 0.9072 & 0.9019 & 0.7673 & 1.0000 & 1.0000 & 1.0000 \\
 &  & Decoder-only & 0.9384 & 0.9042 & 0.7782 & \textbf{0.9623} & \textbf{0.9569} & \textbf{0.9444} \\
 &  & VPT-shallow & 0.9316 & 0.9040 & 0.7951 & 0.8645 & 0.8623 & 0.8568 \\
 &  & VPT-deep & 0.9330 & 0.9037 & 0.7886 & \underline{0.9017} & \underline{0.9026} & \underline{0.8989} \\
 &  & LoRA & 0.9421 & 0.9240 & \textbf{\underline{0.8494}} & 0.6125 & 0.6136 & 0.6064 \\
 &  & LoRA Encoder-only & 0.9436 & 0.9052 & 0.7890 & 0.6511 & 0.6498 & 0.6449 \\
 &  & Full FT & \underline{0.9446} & \textbf{\underline{0.9244}} & 0.8386 & 0.7488 & 0.7425 & 0.7260 \\

\midrule
\multirow{22}{*}{\makecell{PH$^2$ \\ \textit{near-OOD}}} & 0 & Zero-shot & 0.9054 & 0.9004 & 0.7986 & 1.0000 & 1.0000 & 1.0000 \\
 &  & Decoder-only & 0.9466 & 0.8802 & 0.7269 & \textbf{0.9423} & \textbf{0.9385} & \textbf{0.9314} \\
 &  & VPT-shallow & 0.9436 & 0.8846 & 0.7485 & \underline{0.8905} & \underline{0.8958} & \underline{0.8933} \\
 &  & VPT-deep & 0.9458 & 0.8883 & 0.7463 & 0.8825 & 0.8810 & 0.8765 \\
 &  & LoRA & 0.9557 & \underline{0.9053} & \underline{0.8342} & 0.5716 & 0.5766 & 0.5733 \\
 &  & LoRA Encoder-only & \textbf{\underline{0.9598}} & 0.8804 & 0.7528 & 0.6535 & 0.6466 & 0.6331 \\
 &  & Full FT & 0.9583 & 0.8898 & 0.7437 & 0.8258 & 0.8201 & 0.8064 \\
\cmidrule(lr){2-9}
 & 20 & Zero-shot & 0.9054 & 0.9004 & 0.7986 & 1.0000 & 1.0000 & 1.0000 \\
 &  & Decoder-only & 0.9473 & 0.9024 & 0.7753 & \textbf{0.9642} & \textbf{0.9593} & \textbf{0.9492} \\
 &  & VPT-shallow & 0.9421 & 0.8922 & 0.7516 & 0.9004 & 0.8934 & 0.8808 \\
 &  & VPT-deep & 0.9444 & 0.8898 & 0.7465 & \underline{0.9067} & \underline{0.9066} & \underline{0.9013} \\
 &  & LoRA & 0.9516 & \underline{0.9032} & \underline{0.8258} & 0.6391 & 0.6337 & 0.6178 \\
 &  & LoRA Encoder-only & 0.9497 & 0.8750 & 0.7408 & 0.7282 & 0.7234 & 0.7129 \\
 &  & Full FT & \underline{0.9543} & 0.8996 & 0.7533 & 0.8398 & 0.8350 & 0.8251 \\
\cmidrule(lr){2-9}
 & rand100 & Zero-shot & 0.9054 & 0.9004 & 0.7986 & 1.0000 & 1.0000 & 1.0000 \\
 &  & Decoder-only & 0.9408 & 0.9053 & 0.7916 & \textbf{0.9552} & \textbf{0.9512} & \textbf{0.9406} \\
 &  & VPT-shallow & 0.9382 & 0.9106 & 0.8092 & 0.8674 & 0.8665 & 0.8619 \\
 &  & VPT-deep & 0.9368 & 0.9060 & 0.7899 & \underline{0.9114} & \underline{0.9080} & \underline{0.8975} \\
 &  & LoRA & \underline{0.9510} & \textbf{\underline{0.9306}} & \textbf{\underline{0.8724}} & 0.6363 & 0.6349 & 0.6241 \\
 &  & LoRA Encoder-only & 0.9475 & 0.9102 & 0.8332 & 0.6631 & 0.6616 & 0.6544 \\
 &  & Full FT & 0.9468 & 0.9265 & 0.8651 & 0.7936 & 0.7882 & 0.7718 \\

\midrule
\multirow{22}{*}{\makecell{CBIS-DDSM \\ \textit{far-OOD}}} & 0 & Zero-shot & 0.6924 & \textbf{\underline{0.5986}} & \textbf{\underline{0.1889}} & 1.0000 & 1.0000 & 1.0000 \\
 &  & Decoder-only & 0.8273 & 0.5076 & 0.1570 & \underline{0.9353} & \underline{0.9339} & \underline{0.9345} \\
 &  & VPT-shallow & 0.6388 & 0.4264 & 0.1461 & 0.8278 & 0.8361 & 0.8531 \\
 &  & VPT-deep & 0.5744 & 0.3876 & 0.1336 & 0.7996 & 0.8104 & 0.8221 \\
 &  & LoRA & 0.4979 & 0.3795 & 0.1433 & 0.7624 & 0.7710 & 0.7882 \\
 &  & LoRA Encoder-only & 0.7926 & 0.5358 & 0.1872 & 0.9073 & 0.9070 & 0.9092 \\
 &  & Full FT & \textbf{\underline{0.8280}} & 0.5480 & 0.1704 & \textbf{0.9479} & \textbf{0.9487} & \textbf{0.9533} \\
\cmidrule(lr){2-9}
 & 20 & Zero-shot & 0.6924 & \textbf{\underline{0.5986}} & \textbf{\underline{0.1889}} & 1.0000 & 1.0000 & 1.0000 \\
 &  & Decoder-only & 0.7872 & 0.5136 & 0.1565 & \textbf{0.9607} & \textbf{0.9584} & \underline{0.9589} \\
 &  & VPT-shallow & 0.4167 & 0.3157 & 0.1273 & 0.8116 & 0.8285 & 0.8515 \\
 &  & VPT-deep & 0.5767 & 0.3965 & 0.1367 & 0.8324 & 0.8462 & 0.8663 \\
 &  & LoRA & 0.5422 & 0.3951 & 0.1392 & 0.7718 & 0.7857 & 0.8088 \\
 &  & LoRA Encoder-only & 0.7467 & 0.4796 & 0.1628 & 0.8957 & 0.8984 & 0.9078 \\
 &  & Full FT & \underline{0.8015} & 0.5577 & 0.1744 & \underline{0.9547} & \underline{0.9560} & \textbf{0.9613} \\
\cmidrule(lr){2-9}
 & rand100 & Zero-shot & 0.6924 & \textbf{\underline{0.5986}} & \textbf{\underline{0.1889}} & 1.0000 & 1.0000 & 1.0000 \\
 &  & Decoder-only & 0.6527 & 0.4749 & 0.1528 & \underline{0.9115} & \textbf{0.9166} & \textbf{0.9268} \\
 &  & VPT-shallow & 0.2496 & 0.2327 & 0.1101 & 0.7664 & 0.7767 & 0.7917 \\
 &  & VPT-deep & 0.4049 & 0.3178 & 0.1241 & 0.8023 & 0.8172 & 0.8358 \\
 &  & LoRA & 0.1785 & 0.2076 & 0.1073 & 0.7738 & 0.7849 & 0.8003 \\
 &  & LoRA Encoder-only & \underline{0.7410} & 0.4769 & 0.1592 & 0.8850 & 0.8864 & 0.8920 \\
 &  & Full FT & 0.7064 & 0.5159 & 0.1626 & \textbf{0.9152} & \underline{0.9137} & \underline{0.9146} \\

\midrule
\multirow{22}{*}{\makecell{BUSI \\ \textit{far-OOD}}} & 0 & Zero-shot & 0.8234 & \textbf{\underline{0.8156}} & \textbf{\underline{0.6017}} & 1.0000 & 1.0000 & 1.0000 \\
 &  & Decoder-only & 0.8944 & 0.7663 & 0.4888 & \underline{0.9442} & \underline{0.9415} & \underline{0.9370} \\
 &  & VPT-shallow & 0.8112 & 0.7218 & 0.4889 & 0.8599 & 0.8605 & 0.8625 \\
 &  & VPT-deep & 0.7995 & 0.7047 & 0.4695 & 0.8459 & 0.8466 & 0.8458 \\
 &  & LoRA & 0.7775 & 0.6882 & 0.4870 & 0.7888 & 0.7952 & 0.8018 \\
 &  & LoRA Encoder-only & \textbf{\underline{0.9078}} & 0.7732 & 0.5253 & 0.8915 & 0.8887 & 0.8821 \\
 &  & Full FT & 0.8991 & 0.7901 & 0.5014 & \textbf{0.9591} & \textbf{0.9568} & \textbf{0.9528} \\
\cmidrule(lr){2-9}
 & 20 & Zero-shot & 0.8234 & \textbf{\underline{0.8156}} & \textbf{\underline{0.6017}} & 1.0000 & 1.0000 & 1.0000 \\
 &  & Decoder-only & 0.8828 & 0.7963 & 0.5370 & \textbf{0.9628} & \underline{0.9602} & \underline{0.9559} \\
 &  & VPT-shallow & 0.7701 & 0.6944 & 0.4781 & 0.8787 & 0.8787 & 0.8792 \\
 &  & VPT-deep & 0.8136 & 0.7329 & 0.4884 & 0.8935 & 0.8959 & 0.8990 \\
 &  & LoRA & 0.8109 & 0.7362 & 0.5126 & 0.7981 & 0.7965 & 0.7927 \\
 &  & LoRA Encoder-only & \underline{0.8949} & 0.7734 & 0.5227 & 0.8972 & 0.8931 & 0.8864 \\
 &  & Full FT & 0.8935 & 0.8094 & 0.5241 & \underline{0.9611} & \textbf{0.9608} & \textbf{0.9591} \\
\cmidrule(lr){2-9}
 & rand100 & Zero-shot & 0.8234 & \textbf{\underline{0.8156}} & \textbf{\underline{0.6017}} & 1.0000 & 1.0000 & 1.0000 \\
 &  & Decoder-only & 0.8322 & 0.7771 & 0.5114 & \textbf{0.9338} & \textbf{0.9319} & \textbf{0.9291} \\
 &  & VPT-shallow & 0.6723 & 0.6303 & 0.4509 & 0.8193 & 0.8197 & 0.8198 \\
 &  & VPT-deep & 0.7631 & 0.7070 & 0.4747 & 0.8614 & 0.8620 & 0.8604 \\
 &  & LoRA & 0.7474 & 0.7132 & 0.5179 & 0.8039 & 0.8086 & 0.8114 \\
 &  & LoRA Encoder-only & \underline{0.8954} & 0.7722 & 0.5190 & 0.8740 & 0.8744 & 0.8738 \\
 &  & Full FT & 0.8783 & 0.8085 & 0.5135 & \underline{0.9247} & \underline{0.9253} & \underline{0.9223} \\
\bottomrule
\end{longtable}
}

{\small
\begin{longtable}{lllccccc}
\caption{Dice under bounding-box jitter. Models are trained with 0 px, 20 px, or random 0--100 px jitter and evaluated at 0--200 px jitter. Values are means over 3 seeds; underline denotes the best method within each training block and bold the best per dataset/evaluation jitter.}
\label{tab:jitter_all_datasets_dice}\\
\toprule
\textbf{Dataset} & \textbf{Train} & \textbf{Method}
& \textbf{E0} & \textbf{E20} & \textbf{E50}
& \textbf{E100} & \textbf{E200} \\
\midrule
\endfirsthead

\toprule
\textbf{Dataset} & \textbf{Train} & \textbf{Method}
& \textbf{E0} & \textbf{E20} & \textbf{E50}
& \textbf{E100} & \textbf{E200} \\
\midrule
\endhead

\bottomrule
\endfoot

\multirow{21}{*}{\makecell{ISIC 2018 \\ \textit{trained on}}}
& \multirow{7}{*}{pm=0}
& Zero-shot    & 0.9072 & 0.9206 & \underline{0.9144} & \underline{0.8708} & 0.7673 \\
& & Decoder-only & 0.9487 & 0.9349 & 0.8900 & 0.8176 & 0.7102 \\
& & VPT-shallow  & 0.9449 & 0.9327 & 0.8952 & 0.8335 & 0.7352 \\
& & VPT-deep     & 0.9470 & 0.9343 & 0.8944 & 0.8281 & 0.7251 \\
& & LoRA         & 0.9556 & 0.9416 & 0.9040 & 0.8522 & \underline{0.7867} \\
& & LoRA Encoder-only & 0.9580 & 0.9364 & 0.8832 & 0.7951 & 0.6450 \\
& & Full FT      & \textbf{\underline{0.9610}} & \textbf{\underline{0.9483}} & 0.9024 & 0.8259 & 0.7115 \\
\cmidrule(lr){2-8}

& \multirow{7}{*}{pm=20}
& Zero-shot    & 0.9072 & 0.9206 & \underline{0.9144} & \underline{0.8708} & 0.7673 \\
& & Decoder-only & 0.9462 & 0.9365 & 0.9021 & 0.8331 & 0.7271 \\
& & VPT-shallow  & 0.9428 & 0.9326 & 0.9042 & 0.8476 & 0.7478 \\
& & VPT-deep     & 0.9432 & 0.9324 & 0.9014 & 0.8407 & 0.7380 \\
& & LoRA         & 0.9492 & 0.9392 & 0.9114 & 0.8629 & \underline{0.7974} \\
& & LoRA Encoder-only & 0.9496 & 0.9411 & 0.8950 & 0.7885 & 0.6330 \\
& & Full FT      & \underline{0.9556} & \underline{0.9438} & 0.9050 & 0.8314 & 0.7201 \\
\cmidrule(lr){2-8}

& \multirow{7}{*}{rand100}
& Zero-shot    & 0.9072 & 0.9206 & 0.9144 & 0.8708 & 0.7673 \\
& & Decoder-only & 0.9377 & 0.9314 & 0.9145 & 0.8761 & 0.7896 \\
& & VPT-shallow  & 0.9303 & 0.9232 & 0.9075 & 0.8725 & 0.7896 \\
& & VPT-deep     & 0.9343 & 0.9279 & 0.9121 & 0.8766 & 0.7946 \\
& & LoRA         & 0.9411 & 0.9366 & 0.9256 & 0.9018 & \textbf{\underline{0.8478}} \\
& & LoRA Encoder-only & 0.9436 & \underline{0.9429} & \textbf{\underline{0.9371}} &  \textbf{\underline{0.9132}} & 0.7890 \\
& & Full FT      & \underline{0.9445} & 0.9394 & 0.9268 & 0.8979 & 0.8393 \\

\midrule

\multirow{21}{*}{\makecell{PH$^2$ \\ \textit{near-OOD}}}
& \multirow{7}{*}{pm=0}
& Zero-shot    & 0.9054 & 0.9150 & \underline{0.9107} & \underline{0.8756} & 0.7986 \\
& & Decoder-only & 0.9467 & 0.9312 & 0.8917 & 0.8229 & 0.7309 \\
& & VPT-shallow  & 0.9433 & 0.9298 & 0.8976 & 0.8360 & 0.7542 \\
& & VPT-deep     & 0.9456 & 0.9315 & 0.8961 & 0.8324 & 0.7425 \\
& & LoRA         & 0.9570 & 0.9384 & 0.9045 & 0.8600 & \underline{0.8197} \\
& & LoRA Encoder-only & 0.9598 & 0.9402 & 0.9032 & 0.8459 & 0.7528 \\
& & Full FT      & \textbf{\underline{0.9583}} & \underline{0.9385} & 0.8981 & 0.8320 & 0.7434 \\
\cmidrule(lr){2-8}

& \multirow{7}{*}{pm=20}
& Zero-shot    & 0.9054 & 0.9150 & 0.9107 & \underline{0.8756} & 0.7986 \\
& & Decoder-only & 0.9449 & 0.9329 & 0.9019 & 0.8382 & 0.7513 \\
& & VPT-shallow  & 0.9426 & 0.9310 & 0.9040 & 0.8482 & 0.7591 \\
& & VPT-deep     & 0.9428 & 0.9308 & 0.9022 & 0.8431 & 0.7533 \\
& & LoRA         & 0.9511 & 0.9379 & \underline{0.9128} & 0.8717 & \underline{0.8338} \\
& & LoRA Encoder-only & 0.9497 & 0.9363 & 0.9051 & 0.8429 & 0.7408 \\
& & Full FT      & \underline{0.9537} & \underline{0.9398} & 0.9044 & 0.8400 & 0.7516 \\
\cmidrule(lr){2-8}

& \multirow{7}{*}{rand100}
& Zero-shot    & 0.9054 & 0.9150 & 0.9107 & 0.8756 & 0.7986 \\
& & Decoder-only & 0.9390 & 0.9315 & 0.9132 & 0.8711 & 0.7944 \\
& & VPT-shallow  & 0.9359 & 0.9286 & 0.9125 & 0.8736 & 0.7990 \\
& & VPT-deep     & 0.9379 & 0.9314 & 0.9161 & 0.8766 & 0.7988 \\
& & LoRA         & \underline{0.9496} & \textbf{\underline{0.9442}} & \textbf{\underline{0.9329}} & \textbf{\underline{0.9096}} & \textbf{\underline{0.8769}} \\
& & LoRA Encoder-only & 0.9475 & 0.9410 & 0.9280 & 0.9014 & 0.8332 \\
& & Full FT      & 0.9452 & 0.9396 & 0.9290 & 0.9015 & 0.8633 \\

\midrule

\multirow{21}{*}{\makecell{BUSI \\ \textit{far-OOD}}}
& \multirow{7}{*}{pm=0}
& Zero-shot    & 0.8234 & 0.8504 & \underline{0.8384} & \textbf{\underline{0.7579}} & \textbf{\underline{0.6017}} \\
& & Decoder-only & 0.8935 & 0.8693 & 0.7785 & 0.6587 & 0.4915 \\
& & VPT-shallow  & 0.7921 & 0.7758 & 0.7190 & 0.6315 & 0.4868 \\
& & VPT-deep     & 0.8017 & 0.7842 & 0.7243 & 0.6288 & 0.4824 \\
& & LoRA         & 0.7801 & 0.7540 & 0.6975 & 0.6186 & 0.4903 \\
& & LoRA Encoder-only & \textbf{\underline{0.9078}} & \textbf{\underline{0.8983}} & 0.8336 & 0.7012 & 0.5253 \\
& & Full FT      & 0.9007 & 0.8860 & 0.8051 & 0.6722 & 0.4996 \\
\cmidrule(lr){2-8}

& \multirow{7}{*}{pm=20}
& Zero-shot    & 0.8234 & 0.8504 & 0.8384 & \underline{0.7579} & \underline{0.6017} \\
& & Decoder-only & 0.8807 & 0.8717 & 0.8088 & 0.6879 & 0.5184 \\
& & VPT-shallow  & 0.7623 & 0.7509 & 0.7091 & 0.6261 & 0.4790 \\
& & VPT-deep     & 0.7702 & 0.7624 & 0.7224 & 0.6360 & 0.4853 \\
& & LoRA         & 0.7745 & 0.7586 & 0.7162 & 0.6347 & 0.5010 \\
& & LoRA Encoder-only & 0.8949 & \underline{0.8937} & \textbf{\underline{0.8470}} & 0.7089 & 0.5227 \\
& & Full FT      & \underline{0.8996} & 0.8889 & 0.8236 & 0.6902 & 0.5096 \\
\cmidrule(lr){2-8}

& \multirow{7}{*}{rand100}
& Zero-shot    & 0.8234 & 0.8504 & 0.8384 & \underline{0.7579} & \underline{0.6017} \\
& & Decoder-only & 0.8226 & 0.8168 & 0.7883 & 0.6989 & 0.5142 \\
& & VPT-shallow  & 0.6945 & 0.6865 & 0.6652 & 0.6105 & 0.4725 \\
& & VPT-deep     & 0.7360 & 0.7291 & 0.7019 & 0.6343 & 0.4751 \\
& & LoRA         & 0.7467 & 0.7396 & 0.7155 & 0.6542 & 0.4965 \\
& & LoRA Encoder-only & \underline{0.8954} & \underline{0.8807} & 0.8414 & 0.7246 & 0.5190 \\
& & Full FT      & 0.8762 &0.8712 & \underline{0.8424} & 0.7422 & 0.5445 \\

\midrule

\multirow{21}{*}{\makecell{CBIS-DDSM \\ \textit{far-OOD}}}
& \multirow{7}{*}{pm=0}
& Zero-shot    & 0.6924 & 0.7483 & \textbf{\underline{0.6570}} & \textbf{\underline{0.3905}} & \textbf{\underline{0.1889}} \\
& & Decoder-only & 0.8280 & 0.7059 & 0.5039 & 0.3026 & 0.1574 \\
& & VPT-shallow  & 0.5662 & 0.5199 & 0.3961 & 0.2565 & 0.1374 \\
& & VPT-deep     & 0.5725 & 0.5102 & 0.3904 & 0.2582 & 0.1368 \\
& & LoRA         & 0.5094 & 0.4782 & 0.3883 & 0.2703 & 0.1467 \\
& & LoRA Encoder-only & 0.7926 & \textbf{\underline{0.7530}} & 0.5792 & 0.3668 & 0.1872 \\
& & Full FT      & \textbf{\underline{0.8285}} & 0.7438 & 0.5618 & 0.3364 & 0.1713 \\
\cmidrule(lr){2-8}

& \multirow{7}{*}{pm=20}
& Zero-shot    & 0.6924 & \underline{0.7483} & \underline{0.6570} & \underline{0.3905} & \underline{0.1889} \\
& & Decoder-only & 0.7824 & 0.7012 & 0.5186 & 0.3040 & 0.1532 \\
& & VPT-shallow  & 0.4555 & 0.4369 & 0.3532 & 0.2380 & 0.1313 \\
& & VPT-deep     & 0.5117 & 0.4810 & 0.3910 & 0.2603 & 0.1369 \\
& & LoRA         & 0.4370 & 0.4193 & 0.3580 & 0.2472 & 0.1327 \\
& & LoRA Encoder-only & 0.7467 & 0.6743 & 0.4987 & 0.3153 & 0.1628 \\
& & Full FT      & \underline{0.8051} & 0.7398 & 0.5705 & 0.3414 & 0.1714 \\
\cmidrule(lr){2-8}

& \multirow{7}{*}{rand100}
& Zero-shot    & 0.6924 & \underline{0.7483} & \underline{0.6570} & \underline{0.3905} & \underline{0.1889} \\
& & Decoder-only & 0.6090 & 0.5583 & 0.4528 & 0.2764 & 0.1362 \\
& & VPT-shallow  & 0.2759 & 0.2804 & 0.2574 & 0.1956 & 0.1132 \\
& & VPT-deep     & 0.3518 & 0.3349 & 0.2903 & 0.2053 & 0.1091 \\
& & LoRA         & 0.2143 & 0.2244 & 0.2146 & 0.1676 & 0.0949 \\
& & LoRA Encoder-only & \underline{0.7410} & 0.6569 & 0.5074 & 0.3201 & 0.1592 \\
& & Full FT      & 0.6980 & 0.6497 & 0.5380 & 0.3353 & 0.1575 \\

\bottomrule
\end{longtable}
}

\end{document}